\documentclass[11pt]{article}
\usepackage{coling2016}
\usepackage{times}
\usepackage{url}
\usepackage{latexsym}
\usepackage{CJK}
\usepackage{multirow}
\AtBeginDvi{\input{zhwinfonts}}

\title{A Comparison of Word Embeddings for English and Cross-Lingual Chinese Word Sense Disambiguation}

\author{
Hong Jin Kang$^{1}$, Tao Chen$^{2}$,  Muthu Kumar Chandrasekaran$^{1}$, Min-Yen Kan$^{1,2}$\thanks{ This research is supported by the Singapore National Research Foundation
under its International Research Centre @ Singapore Funding
Initiative and administered by the IDM Programme Office.} \\
$^{1}$School of Computing, National University of Singapore\\
$^{2}$NUS Interactive and Digital Media Institute\\
\\
{\tt  \{kanghongjin\}@gmail.com} \\
{\tt  \{taochen,muthu.chandra,kanmy\}@comp.nus.edu.sg} \\
}

\date{June 2016}

\begin{document}
\maketitle

\begin{abstract}
Word embeddings are now ubiquitous forms of word representation in
natural language processing.  There have been applications of 
word embeddings for monolingual word sense disambiguation (WSD) in English,
but few comparisons have been done.  This paper attempts to bridge
that gap by examining popular embeddings for the task of monolingual
English WSD.  Our simplified method leads to comparable
state-of-the-art performance without expensive retraining.

Cross-Lingual WSD -- where the word senses of a word in a source
language $e$ come from a separate target translation language $f$ --
can also assist in language learning; for example, when providing
translations of target vocabulary for learners.  Thus we have also
applied word embeddings to the novel task of cross-lingual WSD for
Chinese and provide a public dataset for further benchmarking.
We have also experimented with using word embeddings for LSTM networks
and found surprisingly that a basic LSTM network does not work well.
We discuss the ramifications of this outcome.
\end{abstract}

\section{Introduction}
\label{intro}

\blfootnote{
    \hspace{-0.65cm}  %
    This work is licensed under a Creative Commons 
    Attribution 4.0 International Licence.
    Licence details:
    \url{http://creativecommons.org/licenses/by/4.0/}
}

A word takes on different meanings, largely dependent on the context
in which it is used. For example, the word ``bank'' could mean ``slope
beside a body of water'', or a ``depository financial
institution''~\footnote{\url{http://wordnetweb.princeton.edu/perl/webwn?s=bank}}. Word
Sense Disambiguation (WSD) is the task of identifying the contextually
appropriate meaning of the word. WSD is often considered a
classification task, in which the classifier predicts the sense from a
possible set of senses, known as a sense inventory, given the target
word and the contextual information of the target word. Existing WSD
systems can be categorised into either data-driven supervised or
knowledge-rich approaches. Both approaches are considered to be
complementary to each other.

Word embeddings have become a popular word representation formalism,
and many tasks can be done using word embeddings. The effectiveness of
using word embeddings has been shown in
several NLP tasks \cite{Turian10wordrepresentations}. The goal of our
work is to apply and comprehensively compare different uses of word
embeddings, solely with respect to WSD. We perform evaluation of the
effectiveness of word embeddings on monolingual WSD tasks from
Senseval-2 (held in 2001), Senseval-3 (held in 2004), and
SemEval-2007. After which, we evaluate our approach on English--Chinese
Cross-Lingual WSD using a dataset that we constructed for %
evaluating our approach on the translation task used in educational
applications for language learning. %

\section{Related Work}

Word Sense Disambiguation is a well-studied problem, in which many
methods have been applied. Existing methods can be broadly categorised
into supervised approaches, where machine learning techniques are used
to learn from labeled training data; and unsupervised %
techniques, which do not rely on labeled data. Unsupervised techniques
are knowledge-rich, and rely heavily on knowledge bases and thesaurus,
such as WordNet~\cite{Miller1995}. It is noted by Navigli \shortcite{Navigli09wordsense}
that supervised approaches using memory-based learning and SVM
approaches have worked best.

Supervised approaches involve the extraction of features and then
classification using machine learning. Zhong and Ng
\shortcite{Zhong2010} developed an open-source WSD system, {\it
  ItMakesSense} (hereafter, IMS), which was considered the
state-of-the-art at the time it was developed.  It is a supervised WSD
system, which had to be trained prior to use. IMS uses three feature
types: 1) individual words in the context surrounding the target word,
2) specific ordered sequences of words appearing at specified offsets
from the target word, and 3) Part-Of-Speech tags of the surrounding three
words.

Each of the features are binary features, and IMS trains a model for
each word. IMS then uses an support vector machine (SVM) for
classification. IMS is open-source, provides state-of-the-art
performance, and is easy to extend. As such, our work features IMS and
extends off of this backbone.

Training data is required to train IMS.  We make use of the
One-Million Sense-Tagged Instances \cite{taghipour2015one} dataset,
which is the largest dataset we know of for training WSD systems, in
training our systems for the All Words tasks.

WSD systems can be evaluated using either fine-grained scoring or
coarse-grained scoring. Under fine-grained scoring, every sense is
equally distinct from each other, and answers must exactly match. 
WordNet is often used as the sense inventory for monolingual WSD tasks. However, WordNet is a fine-grained resource, and even human annotators have
trouble distinguishing between different senses of a word
\cite{edmonds2002introduction}.  In contrast, under coarse-grained
scoring, similar senses are grouped and treated as a single sense.  In
some WSD tasks in SemEval, coarse-grained scoring was done in order
to deal with the problem of reliably distinguishing fine-grained
senses. 

\subsection{Cross-Lingual Word Sense Disambiguation}
Cross-Lingual WSD was, in part, conceived as a further attempt to
solve this issue. In Cross-Lingual WSD, the specificity of a sense is
determined by its correct translation in another language. The sense
inventory is the possible translations of each word in another
language. Two instances are said to have the same sense if they map to
the same translation in that language.
SemEval-2010~\cite{Lefever2010}\footnote{\url{http://stel.ub.edu/semeval2010-coref/}}
and SemEval-2013~\cite{Lefever2013}\footnote{\url{https://www.cs.york.ac.uk/semeval-2013/}}
featured iterations of this task. These tasks featured
English nouns as the source words, and word senses as translations in
Dutch, French, Italian, Spanish and German.

Traditional WSD approaches are used in Cross-Lingual WSD, although
some approaches leverage statistical machine translation (SMT) methods
and features from translation. Cross-Lingual WSD involves training by
making use of parallel or multilingual corpora. In the Cross-Lingual
WSD task in SemEval-2013, the top performing approaches used either
classification or SMT approaches.

\subsection{WSD with Word Embeddings}

In NLP, words can be represented 
with a distributed representation, such as word embeddings, which encodes
words into a low dimensional space. In word embeddings, information
about a word is distributed across multiple dimensions, and similar
words are expected to be close to each other in the vector space. Examples of word
embeddings are Continuous Bag of Words \cite{mikolovword2vec},
Collobert \& Weston's Embeddings \cite{collobert2008unified}, and
GLoVe \cite{pennington2014glove}. We implemented and evaluated the use
of word embedding features using these three embeddings in IMS.

An unsupervised approach using word embeddings for WSD is described by Chen \shortcite{chen2014}. This approach finds representation of senses, instead of words, and computes a context vector which is used during disambiguation. 

A different approach is to work on extending existing WSD
systems. Turian \shortcite{Turian10wordrepresentations} suggests that
for any existing supervised NLP system, a general way of improving
accuracy would be to use unsupervised word representations as
additional features. Taghipour \shortcite{Taghipour15} used C\&W
embeddings as a starting point and implemented word embeddings as a
feature type in IMS. For a specified window, vectors for the
surrounding words in the windows, excluding the target word, are
obtained from the embeddings and are concatenated, producing $d *
(w-1)$ features, where $d$ is the number of dimensions of the vector,
and $w$ is the window size. Each feature is a floating point number,
which is the value of the vector in a dimension. We note that
\cite{Taghipour15} only reported results for C\&W embeddings, and did
not experiment on other types of word embeddings.

Other supervised approaches using word embeddings include AutoExtend
\cite{rothe2015autoextend}, which extended word embeddings to create
embeddings for synsets and lexemes. In their work, they also extended
IMS, but used their own embeddings. The feature types
introduced by this work bear similarities to how Taghipour used
word embeddings, but without Taghipour's method of scaling each
dimension of the word embeddings.

To conclude, word embeddings have been used in several methods to
improve on state-of-the-art results in WSD. However, to date, there
has been little work investigating how different word embeddings and
parameters affect performance of baseline methods of WSD. As far as we
know, there has only been one paper comparing the different word
embeddings with the use of basic composition methods in WSD. Iacobacci
\shortcite{Iacobacci2016} performed an evaluation study of different
parameters when enhancing an existing supervised WSD system with word
embeddings. Iacobacci noted that the integration of Word2Vec
(skip-gram) with IMS was consistently helpful and provided the best
performance. Iacobacci also noted that the composition methods of
average and concatenation produced small gains relative to the other
composition strategies introduced. However, Iacobacci did not
investigate the use of \cite{Taghipour15}'s scaling strategy, which
was crucial to improve the performance of IMS.

We also did not find any recent work attempting to integrate modern
WSD systems for real-world education usage, and to evaluate the WSD
system based on the requirements and suitability for education use.
We aim to fill this gap in applied WSD with this work.

\section{Methods}
\label{section:methods}

As Navigli \shortcite{Navigli09wordsense} noted that supervised approaches have performed best in WSD, we focus on integrating word embeddings in supervised approaches; in specific,
we explore the use of word embeddings within the IMS framework. We focus our work on Continuous Bag of Words (CBOW) from Word2Vec,  Global Vectors for Word Representation (GloVe) and Collobert \& Weston's Embeddings(C\&W). The CBOW embeddings were trained over Wikipedia, while the publicly available vectors from GloVe and C\&W were used. Word2Vec provides 2 architectures for learning word embeddings, Skip-gram and CBOW. In contrast to Iacobacci \shortcite{Iacobacci2016} which focused on Skip-gram, we focused our work on CBOW.
In our first set of evaluations, we used tasks from Senseval-2 (hereafter SE-2), Senseval-3 (hereafter SE-3) and SemEval-2007 (hereafter SE-2007) to evaluate the performance of our classifiers on monolingual WSD. We do this to first validate that our approach is a sound approach of performing WSD, showing improved or identical scores to state-of-the-art systems in most tasks. 

Similar to the work by Taghipour \shortcite{Taghipour15}, we experimented with the use of word embeddings as feature types in IMS. However, we did not just experiment using C\&W embeddings, as different word embeddings are known to vary in quality when evaluated on different tasks \cite{schnabel2015evaluation}. We performed evaluation on several tasks. For the Lexical Sample (LS) tasks of SE-2 \cite{senseval2-LS-kilgarriff2001} and SE-3 \cite{senseval3-LS-mihalcea2004}, we evaluated our system using fine-grained scoring. For the All Words (AW) tasks, fine-grained scoring is done for SE-2 \cite{senseval2-AW-palmer2001} and SE-3 \cite{senseval3-AW-snyder2004}; both the fine \cite{semeval2007-fine-pradhan2007} and coarse-grained were used in \cite{semeval2007-coarse-navigli2007} AW tasks in SE-2007. In order to evaluate our features on the AW task, we trained IMS and the different combinations of features on the One Million Sense-Tagged corpus \cite{taghipour2015one}.

To compose word vectors, one method (used as a baseline) is to sum up
the word vectors of the words in the surrounding context or
sentence. We primarily experimented on this method of composition, due
to its good performance and short training time. For this, every word
vector for every lemma in the sentence (exclusive of the target word)
was summed into a context vector, resulting in $d$ features. Stopwords
and punctuation are discarded. In Turian's
\shortcite{Turian10wordrepresentations} work, two hyperparameters ---
the capacity (number of dimensions) and size of the word embeddings
--- were tuned in his experiments. We follow his protocol and perform
the same in our experiments.

As the remaining features in IMS are binary features, they are not
comparable to the word embeddings which can have unbounded values,
leading to unbalanced influence. As suggested by Turian
\shortcite{Turian10wordrepresentations}, we should scale down the word
embeddings values to the same range as other features. The embeddings are scaled to control their standard
deviations. We implement a variant of this technique as done by
Taghipour \shortcite{Taghipour15}, in which we set the target standard
deviation for each dimension. A comparison of different values of the
scaling parameter, $\sigma$ is done. For each $i \in \{1, 2, .. d\}$:
\\

$E_{i} \leftarrow \sigma \times \frac{E_{i}}{stdev(E_{i})} $, where
$\sigma$ is a scaling constant that sets the target standard deviation
\\

Similar to Turian \shortcite{Turian10wordrepresentations}
and Taghipour \shortcite{Taghipour15}, we found that a value of $0.1$
for $\sigma$ works well, as seen in Table
\ref{table:wordembeddings-accuracy}. 
We evaluate the effect of varying the scaling factor with the feature
of the sum of the surrounding word vectors, and find that the
summation feature works optimally with 50 dimensions.

\begin{table}[th]
	\caption{Effects on accuracy when varying scaling factor on C\&W embeddings}
	\label{table:wordembeddings-accuracy}
	\begin{center}
        \begin{tabular}{| l | r | r |}
			\hline
			{\bf Method} & {\bf SE-2 LS} & {\bf SE-3 LS} \\
			\hline
			C\&W, unscaled & 0.569 & 0.641 \\
			\hline
			C\&W, $\sigma _{=0.15}$ & 0.665 & 0.731 \\
			\hline
			C\&W, $\sigma _{=0.1}$ & {\bf0.672} & {\bf0.739} \\
			\hline
			C\&W, $\sigma _{=0.05}$ & 0.664 & 0.735 \\
			\hline
		\end{tabular}
	\end{center}
\end{table}

In Table \ref{table:top-other-systems}, we evaluate the performance of our system
on both the LS and AW tasks of SE-2 (held in 2001) and SE-3's (held in 2004), and the AW tasks of SE-2007, which were evaluated on by Zhong and Ng \shortcite{Zhong2010}. We obtain statistically significant improvements over IMS on the LS tasks. Our enhancements to IMS to make use of word embeddings also give better results on the AW task than the original IMS, the respective Rank~1 systems from the original shared tasks, and several recent systems developed and evaluated evaluated on the same tasks. We note that although our system increased accuracy on IMS on several
AW tasks, the differences were not statistically significant 
(as measured using McNemar's test for paired nominal data). 

It can be seen that the simple enhancement of integrating word 
embedding using the baseline composition method, followed by 
the scaling step, improves IMS, and we get performance 
comparable to or better than the Rank~1 systems in many tasks.

\begin{table}[th]
	\caption{Comparison of systems by their accuracy score on both Lexical Sample and All Words tasks. Rank 1 system refers to the top ranked system in the respective shared tasks.}
	\label{table:top-other-systems}
	\begin{center}
		\begin{tabular}{|p{4.5cm}|r|r|r|r|r|r|}
			\hline
			{\bf Method} & {\bf SE-2} & {\bf SE-3} & {\bf SE-2} & {\bf SE-3} & {\bf SE-2007} & {\bf SE-2007} \\
             	  &  {\bf LS} &  {\bf LS} & {\bf AW} & {\bf AW} &  {\bf Fine-} & {\bf Coarse-} \\
   	 &	   &     &    &    &  {\bf grained} & {\bf grained} \\
           
			\hline
			IMS + CBOW $\sigma _{=0.1}$ (proposed) & 0.680 & 0.741 & 0.677 & 0.679 & 0.604 & 0.826\\
			\hline
            IMS + CBOW $\sigma _{=0.15}$ (proposed) & 0.670 & 0.734 & 0.673 & 0.675 & {\bf0.615} & {\bf 0.828 } \\
			\hline
			
			IMS & 0.653 & 0.726 & 0.682 & 0.674 & 0.585 & 0.816\\
			\hline
			
			\newcite{rothe2015autoextend} & 0.666 & 0.736 & - & - & - & - \\
			\hline
			\newcite{Taghipour15} & 0.662 & 0.734 & -& {\bf0.682} & - & - \\
			\hline
           	 \cite{Iacobacci2016}  & {\bf0.699} & {\bf0.752} & 0.683 & 0.682 & 0.591 & - \\
            \hline
             \cite{chen2014} & -& - & - & - & - & 0.826  \\
             \hline
			Rank 1 System & 0.642
            & 0.729 
            & {\bf0.69} & 0.652 & 0.591 & 0.825  \\

			\hline
			Baseline (Most Frequent Sense \& Wordnet Sense 1) & 0.476 & 0.552& 0.619 & 0.624 & 0.514 & 0.789 \\
			\hline
		\end{tabular}
	\end{center}
\end{table}

\begin{table}[th]
	\caption{Accuracy of adding word embeddings to IMS, with different parameters, on SE-2, SE-3 LS and AW tasks and SE-2007 AW task}
\vspace{0.15cm}
	\label{table:full}
\centering
\begin{tabular}
{|l|r|r|r|r|r|r|r|r|r|}
\hline
{\bf Type} & {\bf Size} & {\bf Compose} & {\bf Scaling} & {\bf SE-2} & {\bf SE-3} & {\bf SE-2} & {\bf SE-3} & {\bf SE-2007} & {\bf SE-2007} \\
 	&  &  &  &  {\bf LS} &  {\bf LS} & {\bf AW} & {\bf AW} &  {\bf Fine-} & {\bf Coarse-} \\
   	&  &  &  &	   &     &    &    &  {\bf grained} & {\bf grained} \\
\hline
\multirow{3}{*}{C\&W}&\multirow{3}{*}{50}&\multirow{3}{*}{Sum}&0.05&0.666&0.734&0.679&0.673&0.594&0.818 \\

 & & &0.1&0.671&0.738&0.678&0.673&0.6&0.819 \\

 & & &0.15&0.666&0.732&0.675&0.672&0.598&0.817 \\
\hline
\multirow{3}{*}{CBOW}&\multirow{3}{*}{50}&\multirow{3}{*}{Sum}&0.05&0.672&{\bf 0.744}&{\bf 0.68}&0.677&0.604&0.824\\

&&&0.1&{\bf 0.68}&0.741&0.677&0.679 &0.604 & 0.826\\

&&&0.15&0.67&0.734&0.673&0.675&{\bf 0.615}&{\bf 0.828}\\
\hline
\multirow{3}{*}{GloVe}&\multirow{3}{*}{50}&\multirow{3}{*}{Sum}&0.05&0.675&0.738&0.676&0.678&0.596&0.819 \\

& & &0.1&0.679&0.741&0.678&0.68&0.594&0.819 \\

& & &0.15&0.674&0.731&{\bf 0.68}&0.678&0.591&0.819 \\
\hline
\multirow{3}{*}{CBOW}&\multirow{3}{*}{200}&\multirow{3}{*}{Sum}&0.05&0.679&0.742&0.679&0.68&0.602&0.823 \\

& & &0.1&0.669&0.731&0.676&0.675&0.602&0.82 \\

& & &0.15&0.651&0.715&0.667&0.673&0.594&0.822 \\
\hline
\multirow{3}{*}{GloVe}&\multirow{3}{*}{200}&\multirow{3}{*}{Sum}&0.05&0.682&0.741&0.68&{\bf0.682}&0.6&0.823 \\

& & &0.1&0.666&0.73&0.677&0.679&0.591&0.827 \\

& & &0.15&0.654&0.706&0.674&0.675&0.591&0.826 \\
\hline
C\&W&50&Concat&0.1&0.659&0.724&0.679&0.674&0.585&0.818 \\
\hline
\multirow{2}{*}{CBOW}&50&\multirow{2}{*}{Concat}&0.1&0.66&0.725&0.678&0.672&0.581&0.816\\

&200&&0.1&0.667&0.729&0.675&0.67&0.591&0.819\\
\hline
\multirow{2}{*}{GloVe}&50&\multirow{2}{*}{Concat}&0.1&0.657&0.722&0.679&0.671&0.583&0.818\\

&200& &0.1&0.664&0.728&0.677&0.669&0.587&0.817\\
\hline
\end{tabular}
\end{table}

As word embeddings with higher dimensions increases the feature space of IMS, this may lead to overfitting on some datasets. We believe, this is why a
smaller number of dimensions work better in the LS tasks. 
However, as seen in Table~\ref{table:full}, this effect was not
observed in the AW task. We also note that relatively poorer
performance in the LS tasks may not necessarily result in
poor performance in the AW task. We see from the results that
the combination of \cite{Taghipour15}'s scaling strategy and summation
produced results better than the proposal in \cite{Iacobacci2016} to
concatenate and average (0.651 and 0.654), suggesting that the scaling
factor is important for the integration of word embeddings for
supervised WSD.

\subsection{LSTM Network}

A Long Short Term Memory (LSTM) network is a type of Recurrent Neural
Network which has recently been shown to have good performance on many
NLP classification tasks. 
The potential benefit of using an approach using LSTM over our existing
approach in IMS is this is that an LSTM  network is able to use more 
information about the sequence of words. 
For WSD, K{\aa}geb{\"a}ck \& Salomonsson \shortcite{kaageback2016word} explored the use of bidirectional LSTMs. In our approach, we explore a simpler
na\"{\i}ve approach instead.

For the Lexical Sample tasks, we train the model on the training data
provided for the task. For the All Words task, we trained the model on
the One Million Sense-Tagged dataset. For each task, similar to IMS,
we train a model for each word, using GloVe word embeddings as the input layer.

\begin{table}[th]
	\caption{Accuracy of a basic LSTM approach on the Lexical
          Sample and All Words tasks.}
	\label{table:NN-LS}
	\begin{center}
		\begin{tabular}{| p{6cm} | r | r | r | r |}
			\hline
			\textbf{Method} & \textbf{SE-2 LS}  & \textbf{SE-3 LS} & \textbf{SE-2 AW}  & \textbf{SE-3 AW} \\
			\hline
			LSTM approach (Proposed) & 0.458  & 0.603 & 0.619 & 0.623 \\
			\hline
			IMS & 0.653 & 0.726 & 0.682 & 0.674 \\
            \hline
            \cite{kaageback2016word} & 0.669 & 0.734 & - & - \\
			\hline
			Rank 1 System during the task & 0.642 & 0.729 & 0.69 & 0.652 \\
			\hline
			Baseline & 0.476 & 0.552 & 0.619 & 0.624 \\
			\hline
		\end{tabular}
	\end{center}
\end{table}

The performance of the na\"{\i}ve LSTM is poor in both type of tasks, as seen in Table
\ref{table:NN-LS}. The models converge to just
using the most common sense for the AW task. A possible reason for this is overfitting. WSD is known to suffer
from data sparsity. 
Although there are many training
examples in total, as we train a separate model for each word, many
individual words only have few training examples. 
We note other attempts to use neural networks for WSD may have run into the same problem. Taghipour and Ng \shortcite{Taghipour15} indicated the need to prevent overfitting while using a neural network to adapt C\&W embeddings by omitting a hidden layer and adding a Dropout layer, while K{\aa}geb{\"a}ck and Salomonsson \shortcite{kaageback2016word} developed a new regularization technique in their work.

\section{English-Chinese Cross-Lingual Word Sense Disambiguation}
\label{section:CLWSD}

We now evaluate our proposal on the Cross-Lingual Word Sense
Disambiguation task.  One key application of such task is to
facilitate language learning systems.  For example, {\it
 MindTheWord}\footnote{\url{https://chrome.google.com/webstore/detail/mindtheword/fabjlaokbhaoehejcoblhahcekmogbom}}
and {\it WordNews}~\cite{tao2014} are two applications that allow
users to learn vocabulary of a second language in context, in the form
of providing translations of words in an online article.
In this work, we model this problem of finding translations of words
as a variant of WSD, Cross-Lingual Word Sense Disambiguation, as
formalized in \cite{tao2014}.

In the previous section, we have validated and compared enhancements to
IMS using word embeddings. These have produced results comparable to,
and in some cases, better than state-of-the-art performance on the
monolingual WSD tasks. We further evaluate our approach for use in the
Cross-Lingual Word Sense Disambiguation for performing contextually
appropriate translations of single words. To accomplish this, we first
construct a English--Chinese Cross-Lingual WSD dataset. For our sense
inventory, we work with the existing dictionary in the open-source
educational application, WordNews \cite{tao2014}, which contains a
dictionary of English words and their possible Chinese
translations. We finally deploy the trained system as a fork of the
original WordNews.

\subsection{Dataset}
 As far as we know, there 
 is no existing publicly available English--Chinese Cross-Lingual WSD dataset.
 To evaluate our proposal, therefore, we hired human annotators to construct
 such an evaluation dataset using sentences from recent news articles. As the dataset is 
 constructed using recent news data, it is a good representation for the use case in WordNews. To facilitate future research, we have released the dataset to the public.{\footnote{{\url{  https://kanghj.github.io/english_chinese_news_clwsd_dataset/}}}}

To obtain the gold standard for this data set, we hired 18 annotators to select the right translations for a given word and its context. There are 697 instances in total in our dataset, with a total of 251 target words to disambiguate, that were each multiply-annotated by 3 different annotators. Each annotator disambiguated 110+ instances (15 annotators with 116 instances, 3 with 117) in hard-copy. The annotators are all bilingual undergraduate students, who are native Chinese speakers. 

For each instance, which contains a single English target word to disambiguate, we include the sentence it appears in and its adjacent sentences as its context. Each instance contains possible translations of the word. 
The annotators selected all Chinese words that had an identical meaning to the English target word. If the word cannot be appropriately translated, we instructed annotators to leave the annotation blank. The annotators provided their own translations if they believe that there is a suitable translation, but which was not provided by the crawled dictionary. 

The concept of a sense is a human construct, and therefore, as earlier elaborated on when discussing sense granularity, it is %
may be difficult for human annotators to agree on the correct answer. 
Our annotation task differs from the usual since we allow users to select multiple labels and can also add new labels to each case if they do not agree with any label provided. As such, applying the Cohen's Kappa as it is for measuring the inter-annotator agreement as it is does not work for our annotated dataset. %
We are also unable to compute the probably of chance agreement by word, since there are few test instances per word in our dataset.

The Kappa equation is given as 
$\kappa = \frac{p_A - p_E}{1 - p_E} $.
To compute $p_A$ for $\kappa$, we use a simplified, optimistic approach where we select one annotated label out of possibly multiple selected labels for each annotator. We always choose the label that results in an agreement between the pair, if such a label exist. For $p_E$ (the probability of chance agreement), as the labels of each case are different, we consider the labels in terms of how frequent they occur in the training data. 
We only consider the top 3 most frequent senses for each word %
due to the skewness of the sense distribution. 
We first compute the probability of an annotator selecting each of the top three frequent senses, $p_E$ is then equals to the sum of the probability that both annotators selected one of the three top senses by chance. 

The pairwise value by this proposed method of $\kappa$ is obtained is 0.42. We interpreted this as a moderate level of agreement. We note that there is a large number of possible labels for each case, which is known to affect the value of $\kappa$ negatively. This is exacerbated as we allow the annotators to add new labels. 

In this annotation task, as we consider the possible translations as  fine-grained, the value of agreement is likely to be underestimated in this case. Hence, we believe that clustering of similar translations during annotation is required in order to deal with the issue of sense granularity in Cross-Lingual WSD. 
To overcome this problem, we used different configurations of granularity during evaluation of our system. 
For all configurations,
we remove instances from the dataset if it does not have a correct sense. 

We also noticed that some target words were part of a proper noun, such as the word 'white' in 'White House'. This led to some confusion among annotators, so we omitted instances where the target word is part of a proper noun. Statistics of the test dataset after filtering out different cases are given in Table \ref{table:CLWSD-test-stats-no-ne}.

\begin{table}[ht]
	\caption{Statistics of our new annotated Chinese-English crosslingual WSD dataset. Out-of-vocabulary (OOV) annotations refer to annotations added by the annotators}
	\label{table:CLWSD-test-stats-no-ne}
	\begin{center}
		\begin{tabular}{| p{8cm} | r| r|}
			\hline
			{\bf Configuration} & {\bf \# of instances} & {\bf \# of unique target words} \\
			\hline
			Include all & 653 & 251\\ 
			\hline
			Exclude instances with OOV annotations & 481 & 206 \\						
			\hline
			Exclude instances without at least partial agreement & 412 & 193 \\
			\hline
			Exclude instances without complete agreement & 229 & 136 \\
			\hline
		\end{tabular}
	\end{center}
\end{table}

\subsection{Experiments}

As previously described, IMS is a supervised system requiring training data before use. We constructed data by processing a parallel corpus, the news section of the UM-Corpus \cite{tian2014corpus}, and performing word alignment. We used the dictionary provided by \cite{tao2014} as the sense inventory, which we further expanded using translations from Bing Translator and Google Translate. For construction of the training dataset, word alignment is used to assign Chinese words as training labels for each English target word. GIZA++ \cite {och03} is used for word alignment. To evaluate our system, we compare the results of the method described in \cite{tao2014}, which uses Bing Translator and word alignment to obtain translations. We use the configuration where every annotation is considered to be correct for our main evaluation since this is closer to a coarse-grained evaluation. 

\begin{table}[ht]
	\caption{Results of our systems on the Cross-Lingual WSD dataset, excluding named entities. Instances with out-of-vocabulary annotations are removed. All annotations are considered correct answers.}
	\label{table:CLWSD-test-results}
	\begin{center}

			\begin{tabular}{| p{9cm}| r| }
				\hline
				\textbf{Method} & \textbf{Accuracy} \\
				\hline
				Bing Translator + 
                word alignment (baseline) & 0.559  \\
				\hline
				IMS & 0.752  \\
				\hline
                IMS + CBOW, 50 dimensions, $\sigma _{=0.05}$ (proposed) &  0.763  \\
				\hline
				IMS + CBOW, 50 dimensions, $\sigma _{=0.1}$ (proposed) & {\bf 0.772}  \\                                
                \hline
                IMS + CBOW, 50 dimensions, $\sigma _{=0.15}$ (proposed) & 0.767  \\
                \hline
			\end{tabular}

	\end{center}
\end{table}

It can be seen that word embeddings improves the performance on Cross-Lingual WSD. Similar to our observations for monolingual WSD, the use of both CBOW and GLoVe improved performance. However, the improvements from the word embeddings feature type over IMS was not statistically significant at 95\% confidence level. This is attributed to the small size of the dataset. 

\subsection{Bing Translator results}
We wish to highlight and explain the poor performance of Bing Translator with our annotated dataset as seen in Table \ref{table:CLWSD-test-results}. This could be because Bing Translator performs translation at the phrase level. Therefore, many of the target words are not translated individually and are translated only as part of a larger unit, making it less suitable for the use case in WordNews where only the translation of single words matter. 
\begin{CJK*}{UTF8}{gbsn}
For example, when translating the word ``little'' in ``These are serious issues and themes, and sometimes {\bf little} kids aren't ready to process and understand these ideas'', Bing Translator provides a translation of ``这些都是严重的问题和主题，有时{\bf 小孩} 不准备处理和理解这些想法'' but does not give an alignment for the word `little' but instead provides an alignment for the entire multi-word unit ``little kids''. 
\end{CJK*}
As such, the translation would not match any of the annotations provided by our annotators. This is an appropriate treatment since a user of an educational app requesting specifically a translation for the single word ``little'' should not see the translation of the phrase.

\section{Conclusion}
\label{section:conclusion}

After we have evaluated the performance of the systems on the this
Cross-Lingual WSD dataset, we integrate the top-performing system
using word embeddings and the trained models into a fork of the
WordNews system. We experimented and implemented with different
methods of using word embeddings for supervised WSD. We tried two
approaches, by enhancing an existing WSD system, IMS, and by trying a
neural approach using a simple LSTM.  We evaluated our apporach as
well as various methods in WSD, against initial evaluations on the
existing test data sets from Senseval-2, Senseval-3, SemEval-2007. In
a nutshell, adding any pretrained word embedding as a feature type to
IMS resulted in the system performing competitively or better than the
state-of-the-art systems on many of the tasks. This supports
\cite{Iacobacci2016}'s conclusion that concluded that existing
supervised approaches can be augmented with word embeddings to give
better results.

Our findings also validated Iacobacci et al. \shortcite{Iacobacci2016}'s
findings that Word2Vec gave the best performance. However, we also
note that, other than Word2Vec, other publicly available word
embeddings, Collobert \& Weston's embeddings and GLoVe also
consistently enhanced the performance of IMS using the summation
feature with little effort. Other than on the Lexical Sample tasks,
where smaller word embeddings performed better, we also found that the
number of dimensions did not affect results as much as the scaling
parameter. Unlike Iacobacci et al. \shortcite{Iacobacci2016}, we also
found that a simple composition method using summation already gave
good improvements over the standard WSD features, provided that the
scaling method described in \cite{Taghipour15} was performed.

An additional key contribution of our work was to build a
gold-standard English-Chinese Cross-Lingual WSD dataset constructed
with sentences from real news articles and to evaluate our proposed
word embedding approach under this scenario.  Our compiled dataset was
used as evaluation of the task of translating English words on online
news articles. This dataset is made available publicly.  We observed
that word embeddings also improves the performance of WSD in our
Cross-Lingual WSD setting.

As future work, we will examine how to expand the existing dictionary
with more English words of varying difficulty and include more
possible Chinese translations, as we note that there were several
instances in the Cross-Lingual WSD dataset where the annotators did
not choose an existing translation.

\bibliographystyle{acl}
\bibliography{socreport}
\end{document}